\documentclass{article}
\usepackage{spconf,amsmath,graphicx,hyperref}
\usepackage{algorithm}
\usepackage{algpseudocode}
\usepackage{bm}
\usepackage{amsmath}
\usepackage{amssymb}
\usepackage{graphicx}
\usepackage{subcaption}
\usepackage{multirow}
\usepackage{amssymb}
\usepackage{pifont}

\begin{document}
\title{
\vspace{-0.32in}
HPTune: Hierarchical Proactive Tuning for\\Collision-Free Model Predictive Control
\vspace{-0.12in}
}

\name{
\begin{tabular}{c}
    Wei Zuo$^{1}$, Chengyang Li$^{1}$, Yikun Wang$^{1}$, Bingyang Cheng$^{1}$, Zeyi Ren$^{1}$, \\
    Shuai Wang$^{2,\dagger}$, Derrick Wing Kwan Ng$^{3}$, Yik-Chung Wu$^{1,\dagger}$
\end{tabular}
\vspace{-0.12in}
\noindent\thanks{$^\dagger$Corresponding authors: Shuai Wang ({\tt\small s.wang@siat.ac.cn}) and Yik-Chung Wu ({\tt\small ycwu@eee.hku.hk}). 
This work is supported by the National Key R\&D Program of China (No. 2025YFE0204100), Science and Technology Development Fund of Macao S.A.R (FDCT) under number 0074/2025/AMJ,
the National Natural Science Foundation of China (Grant No. 62371444), and the Shenzhen Science and Technology Program (Grant No. RCYX20231211090206005, JCYJ20241202124934046).
}
}
\address{$^{1}$ The University of Hong Kong, \:{Hong Kong} \\ $^{2}$ Shenzhen Institutes of Advanced Technology, Chinese Academy of Sciences,\:{China}\\ $^{3}$ University of New South Wales,\:{Australia}
\vspace{-0.2in}
}  

\maketitle

\begin{abstract}
Parameter tuning is a powerful approach to enhance adaptability in model predictive control (MPC) motion planners.
However, existing methods typically operate in a myopic fashion that only evaluates executed actions, leading to inefficient parameter updates due to the sparsity of failure events (e.g., obstacle nearness or collision).
To cope with this issue, we propose to extend evaluation from executed to non-executed actions, yielding a hierarchical proactive tuning (HPTune) framework that combines both a fast-level tuning and a slow-level tuning. 
The fast one adopts risk indicators of predictive closing speed and predictive proximity distance, and the slow one leverages an extended evaluation loss for closed-loop backpropagation.
Additionally, we integrate HPTune with the Doppler LiDAR that provides obstacle velocities apart from position-only measurements for enhanced motion predictions, thus facilitating the implementation of HPTune.
Extensive experiments on high-fidelity simulator demonstrate that HPTune achieves efficient MPC tuning and outperforms various baseline schemes in complex environments. 
It is found that HPTune enables situation-tailored motion planning by formulating a safe, agile collision avoidance strategy.

\end{abstract}
\section{Introduction} \label{intro}
\vspace{-0.07in}

Motion planning techniques are gaining prominence due to their wide applications, such as emergency rescue and autonomous logistics \cite{han2025hierarchically}.
The core objective of motion planning is to achieve real-time and collision-free navigation for given tasks \cite{teng2023motion}, where model predictive control (MPC) is a commonly adopted control framework \cite{ji2016path}.
The operation of MPC requires appropriate tuning of a cost function to penalize objective deviations, and a series of system constraints \cite{mayne2000constrained}. 
Since manual tunings are often imperfect and inefficient, many studies focus on MPC auto-tuning, which transforms the tuning problem into a parameter optimization problem and updates parameters automatically for better performance \cite{cheng2024difftune}.

Existing auto-tuning methods can be broadly divided into two categories, i.e., open-loop and closed-loop. Open-loop methods tune parameters by directly allowing MPC cost/constraint adaptability inside the prior problem\cite{han2023rda}, or by evaluating the solution with a loss instantly after the problem is solved \cite{amos2018differentiable, jin2021safe}. However, they neglect interactive environmental impacts on the actual execution. In contrast, closed-loop methods use a posterior loss to evaluate system behaviors upon the actual execution, leading to practical update strategies \cite{edwards2021automatic,song2022policy,tao2024difftune}. 
Nevertheless, these methods typically operate in a myopic fashion, because the posterior loss only evaluates current and past states resulting from executed actions, leaving non-executed actions along the MPC horizon unevaluated.
Further due to the sparsity of failure events (e.g., obstacle nearness or collision), such closed-loop tuning suffers from inefficiency.

To fill this gap, this paper proposes hierarchical proactive tuning (HPTune), a closed-loop solution that leverages motion predictions to extend the evaluation from executed to non-executed actions, and benefits from a hierarchical structure that combines both fast-level and slow-level tunings to ensure dense parameter updates.
Specifically, the fast-level tuning allows high-frequency parameter update based on two risk indicators, i.e., the predicted closing speed and the predicted proximity distance. 
The slow-level tuning is executed at a lower frequency, using an extended closed-loop evaluation loss for gradient backpropagation.
We also integrate HPTune with the Doppler light detection and ranging (LiDAR) sensor, a 4D sensor that provides extra velocity measurements besides ranging information in existing 3D LiDAR, and fuse the new velocity dimension with a Kalman filter for enhanced predictions. As such, HPTune leads to a paradigm shift from \textit{reactive closed-loop} towards \textit{proactive closed-loop} tuning.

Finally, we deploy the entire HPTune in the high-fidelity simulator CARLA \cite{dosovitskiy2017carla}, and conduct extensive experiments comparing existing DiffTune-MPC \cite{tao2024difftune}, RDA \cite{han2023rda} and OBCA \cite{zhang2020optimization} methods.
Results show that HPTune increases navigation pass rate by at least $8.3\,\%$ compared to other methods, and excels in diverse metrics including navigation time, motion smoothness, and tuning efficiency.  

\vspace{-0.06in}
\section{Problem Formulation}\label{section2}
\vspace{-0.06in}

\begin{figure}[t]
\centering
\includegraphics[width=0.49\textwidth]{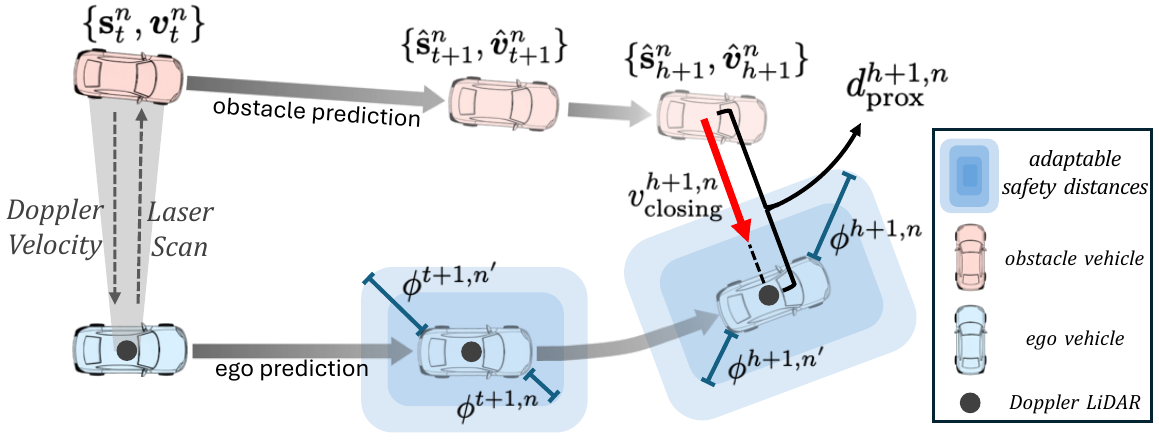}
\caption{Extended evaluation built upon risk indicators. }
\label{indicator}
\vspace{-0.15in}
\end{figure}

Consider a 4-wheel vehicle that performs MPC motion planning among $N$ obstacles, where $n\in\mathcal{N}=\{1,2,\cdots,N\}$ is the obstacle index.
It operates with an $H$-step discrete horizon $h\in\mathcal{H}_t=\{t,\cdots,t+H-1\}$ to determine an optimal action sequence $\mathbf{w}_{h}\in\mathcal{W}_t\!=\!\{\mathbf{w}_t,\mathbf{w}_{t+1},\cdots,\mathbf{w}_{t+H-1}\}$ and the corresponding state sequence $\mathbf{s}_{h}\in\mathcal{S}_t=\{\mathbf{s}_{t+1},\mathbf{s}_{t+2}, \cdots, \mathbf{s}_{t+H}\}$ at time step $t$, while only executing $\mathbf{w}_t$. 
Specifically, $\mathbf{s}_{h}=(\mathbf{z}_h,\theta_h)$, where $\mathbf{z}_h$ and $\theta_h$ are position and orientation; $\mathbf{w}_{h}=(\bm{v}_h,\psi_{h})$, where $\bm{v}_h$ and $\psi_{h}$ are linear and angular velocity. 
To ensure collision avoidance, a safety distance constraint is introduced as:
\vspace{-0.02in}
\begin{align}
    \mathbf{dist}\big(\mathbb{G}(\mathbf{s}_t),\mathbb{O}(\mathbf{s}^n_t)\big)\geq \phi_\text{safe},\label{safty}
\end{align}
where $\phi_\text{safe}$ is the minimal safety distance, $\mathbb{G}(\cdot)$ is the ego bounding box, and $\mathbb{O}(\cdot)$ is the obstacle bounding box. $\mathbf{s}^n_{h}\!=\!(\mathbf{z}^n_h,\theta_h^n)$ is the obstacle state where $\mathbf{z}^n_h$ and $\theta^n_h$ are position and orientation, respectively.
$\mathbf{dist}(\cdot)$ is the minimum-distance operator that calculates the minimal distance between two bounding boxes following \cite{han2025neupan}.
Therefore, the MPC motion planning problem can be formulated as:
\vspace{-0.02in}
\begin{subequations}\label{P}
\begin{align}
    & \mathsf{P}_t(\Gamma):~\min_{\mathcal{S}_t, \mathcal{W}_t}\ C_t \\
    & \text{s.t.}~~~~\mathbf{s}_{h+1} = \mathbf{s}_{h} + f(\mathbf{s}_{h}, \mathbf{w}_{h}) \Delta t, \label{sta_tran}  \\
    & ~~~~~~~~~\mathbf{dist}\big(\mathbb{G}(\mathbf{s}_t),\mathbb{O}(\mathbf{s}^n_t)\big)\geq \phi_\text{safe}, \\
    & ~~~~~~~~~h \in \mathcal{H}_t, \mathbf{w}_h\in \mathbb{W}, n\in\mathcal{N},
\end{align}
\end{subequations}
where $\Delta t$ is the step interval, $\mathbb{W}$ is the action bounds, $f(\cdot)$ is the kinematic maneuvering model following \cite{han2023rda} and $\Gamma$ is the tunable parameter set that can be updated by a posterior evaluation loss $L(t)$.
$C_t$ is the navigation cost defined as:
\vspace{-0.05in}
\begin{align}
    C_t\!=\!\sum_{h \in \mathcal{H}_t} \!\big(
    \alpha\left\| \mathbf{s}_{h+1} \!-\! \mathbf{s}_{h+1}^{\star} \right\|_2^2 \!+\!
    (1\!-\!\alpha)\left\| \mathbf{w}_{h} \!-\! \mathbf{w}_{h}^{\star} \right\|_2^2 \big), \label{ct}
\end{align}
where $\mathbf{s}_{h+1}^{\star}$ and $\mathbf{w}_{h}^{\star}$ are reference navigation tracking state and action, and $\alpha\in(0,1)$ is a tunable weighting factor.
 
\vspace{-0.06in}
\section{Proposed HPTune Method}\label{section3}
\vspace{-0.06in}

In this section, we propose HPTune, which extends the closed-loop evaluation from executed to non-executed actions, constituting a hierarchical proactive tuning framework.
\vspace{-0.12in}

\subsection{Fast-level Parameter Tuning}
\vspace{-0.02in}

This level represents the lower architecture of HPTune, aiming at enabling high-frequency lightweight parameter update.
In an MPC system, although most actions are non-executed, they represent an expectation of optimal future states. Thus, when it comes to the next planning round,
they can be leveraged to serve as an ego prediction sequence over a truncated horizon $\mathcal{H}^-_{t}\!=\!\{t,t+1,\cdots\!,t+H\!-\!2\}$.
Correspondingly, given the obstacle prediction $\{\hat{\mathbf{s}}^n_{h+1}, \hat{\bm{v}}_{h+1}^n\}_{h\in\mathcal{H}^-_t}$ where $\hat{\mathbf{s}}^n_{h+1}=(\hat{\mathbf{z}}^n_{h+1},\hat{\theta}_{h+1}^n)$,
two risk indicators can be formulated:
\vspace{-0.03in}
\begin{itemize}
    \item \textbf{Predictive proximity distance.}
    \begin{align}
    d^{h+1,n}_\text{prox}=
    \mathbf{dist}\big(\mathbb{G}(\mathbf{s}_{h+1}),\mathbb{O}(\hat{\mathbf{s}}^n_{h+1})\big), \forall h\in\mathcal{H}^-_{t},
    \end{align}
    which measures predictive distances between bounding boxes of the ego prediction and the obstacle prediction.
    \vspace{-0.05in}
    \item  \textbf{Predictive closing speed.}
    \begin{align}
        v^{h+1,n}_\text{closing}\! =\! \frac{(\! \bm{v}_{h+1}\!-\!\hat{\bm{v}}^n_{h+1}\!)\! \cdot \!(\hat{\mathbf{z}}^n_{h+1}\! - \!\mathbf{z}_{h+1})}{\|\hat{\mathbf{z}}^n_{h+1}\! -\! \mathbf{z}_{h+1}\|_2}, \forall h\! \in \!\mathcal{H}^-_{t}\!,
    \end{align}
    which is the projection of the relative velocity onto the line connecting ego and obstacle positions, therefore indicating the obstacle incoming speed.
\end{itemize}
\vspace{-0.05in}

As Fig. \ref{indicator} shows, these indicators effectively extend the evaluation to non-executed actions, enabling proactive risk-aware updates. 
Since a constant $d_\text{safe}$ is unadaptable to varying environments and can even result in infeasible $\mathsf{P}_t(\Gamma)$,
we transform $d_\text{safe}$ into obstacle-horizon indexed $\phi^{h+1,n}$ that allows situation-tailored tuning following \cite{han2023rda,dixit2023adaptive}.
Furthermore, $\phi^{h+1,n}$ is decomposed into a base component $\phi_\text{base}$, and a proactive component $\phi_\text{proactive}^{h+1,n}$ which is updated as follows:
\vspace{-0.07in}
\begin{align}\label{beta}
    \phi_\text{proactive}^{h+1,n}\!=\!(\phi_\text{max}\!-\!\phi_\text{base})\tanh\left(\beta\;\text{ReLU}\left (\frac{v^{h+1,n}_\text{closing}}{d^{h+1,n}_\text{prox}}\right)\right),
\end{align}
where $\phi_\text{max}$ is the upper bound of $\phi^{h+1,n}$ and $\beta$ is a tunable sensitivity factor.
The ratio $v^{h+1,n}_\text{closing}/d^{h+1,n}_\text{prox}$ explicitly quantifies the imminence of the potential nearness or collision.
The $\tanh$ activation ensures the smoothness of $\phi_\text{proactive}^{h+1,n}$, while the $\text{ReLU}$ activation ensures that only approaching obstacles contributes.
Consequently, \eqref{P} can be reformulated as
\begin{subequations}
\vspace{-0.05in}
\begin{align}
    & \mathsf{Q}_t(\Gamma):~\min_{\mathcal{S}_t,\mathcal{W}_t}\ C_t \\
    & \text{s.t.}~~~~~~~\mathbf{s}_{h+1} = \mathbf{s}_{h} + f(\mathbf{s}_{h}, \mathbf{w}_{h}) \Delta t,\\
    & ~~~~~~~~~~~~\mathbf{dist}\big(\mathbb{G}(\mathbf{s}_{h+1}),\mathbb{O}(\hat{\mathbf{s}}^n_{h+1})\big)\geq  \phi_\text{base}+\phi_\text{proactive}^{h+1,n}, \\
    & ~~~~~~~~~~~~h \in \mathcal{H}_t, \mathbf{w}_h\in \mathbb{W}, n\in\mathcal{N}.
\end{align}
\label{Q}
\vspace{-0.25
in}
\end{subequations}

To address the challenge of obtaining accurate obstacle predictions, we incorporate the powerful Doppler LiDAR, which features providing Doppler velocity measurement that represents the laser-direction component of the object relative velocity \cite{hexsel2022dicp} and has been adopted in areas including object detection and motion tracking \cite{peng2021detection,gu2022learning,zhao2024fmcw}. Considering that a Doppler LiDAR is mounted on the ego vehicle to scan the surrounding environment, the linear velocity of the $n$-th obstacle is estimated by: 
\vspace{-0.03in}
\begin{align}
     \hat{\bm{v}}_t^n=\frac{1}{I^n}\sum_{i^n\in\mathcal{I}^n}
    \bm{v}_{t,i^n}^\text{Doppler}\mathbf{P}_{t,i^n},
\end{align}
where $i^n\in\mathcal{I}^n$ is the index of $I^n$ points within the $n$-th obstacle bounding box, and $\mathbf{P}_{t,i^n}$ is the projection matrix that projects each Doppler velocity measurement back to corresponding obstacle linear velocity direction following \cite{hexsel2022dicp}. Subsequently, a Kalman filter is adopted to predict obstacle future states along $\mathcal{H}^-_t$, resulting in $\{\hat{\mathbf{s}}^n_{h+1}, \hat{\bm{v}}_{h+1}^n\}_{h\in\mathcal{H}^-_t}$.

\vspace{-0.1in}
\subsection{Slow-level Parameter Tuning}
\vspace{-0.03in}

This level represents the upper architecture of HPTune, aiming at enabling low-frequency backpropagation parameter update.
We select two key factors $\{\alpha,\beta\}$ for $\Gamma$, where $\alpha$ balances navigation tracking and control effort while $\beta$ modulates the sensitivity of proactive safety margins. The extended evaluation loss functions are formulated as follows:
\vspace{-0.03in}
\begin{equation}
    \left\{
    \begin{aligned}
        L_1(t)&\!=\!\sum_{k=t-T+1}^{t}\,\left(\alpha\| \mathbf{s}_{k}\!-\!\mathbf{s}_{k}^{\star}\|_2^2\!+\!(1\!-\!\alpha)\| \mathbf{w}_{k\!-\!1}\!-\!\mathbf{w}_{k\!-\!1}^{\star}\|_2^2\right), \nonumber\\
        L_2(t)&=-\sum_{\substack{n \in \mathcal{N},h \in \mathcal{H}_{t}^-}}\!\mathbf{ReLU}\left(\frac{\phi^{h+1,n}(\beta)-\!d^{h+1,n}_\text{prox}}{d^{h+1,n}_\text{prox}}\right), \nonumber \\
        L_3(t)&=\,\sum_{\substack{n \in \mathcal{N},h \in \mathcal{H}_{t}}}\|\phi^{h+1,n}(\beta)\|_1, \nonumber
    \end{aligned}
    \right.
\end{equation}
where $L_1(t)$ is the navigation loss over the past $T$ executed steps following \cite{tao2024difftune,han2025neupan}; $L_2(t)$ is the predictive safety loss that penalizes if $d^{h+1,n}_\text{prox}$ falls below $\phi^{h+1,n}$; $L_3(t)$ is the $\ell_1$-regularization of safety distances.
Note that the function $\phi^{h+1,n}(\beta)$ is defined in \eqref{beta}.

Together, these loss functions combine to balance among safety, adaptability, and navigation tracking: $L_1(t)$ penalizes tracking violations on executed actions; $L_2(t)$ increases $\phi^{h+1,n}$ for a particular step and obstacle where a potential nearness or collision induced by non-executed actions occurs; $L_3(t)$ penalizes excessive safety margins to ensure feasibility. The overall extended closed-loop evaluation loss is:
\vspace{-0.03in}
\begin{align}
    L(t) =
    \eta_1L_1(t)+ 
    \eta_2L_2(t)+
    \eta_3L_3(t),
\end{align}
where $\eta_1$, $\eta_2$ and $\eta_3$ are weighting coefficients. 
The parameter update is performed every $T$ steps to match the evaluation length of $L_1(t)$, using backpropagation as follows:
\vspace{-0.02in}
\begin{align}
    \gamma\leftarrow\gamma-\epsilon\frac{\partial L(t)}{\partial\gamma},
    \label{update}
\end{align}
where $\gamma\in\Gamma$ is any tunable parameter, $\epsilon$ is the update rate. 
The entire workflow of HPTune is summarized in Alg. \ref{alg:DCTune}, where (\ref{Q}) is solved using duality optimization following \cite{han2023rda}.

\begin{algorithm}[t]
\caption{Workflow of HPTune}
\label{alg:DCTune}
\begin{algorithmic}[1]
\State Initialize $\{\mathcal{S}_{0}, \mathcal{W}_{0}\} = \emptyset$
\State Initialize $\Gamma$
\For{$t = 1, 2, \ldots$}
    \For{$n \in \mathcal{N}$}
        \State Estimate $\hat{\bm{v}}_t^n$ from Doppler LiDAR
        \State Predict $\{\hat{\mathbf{s}}^n_{h+1}, \hat{\bm{v}}_{h+1}^n\}_{h\in\mathcal{H}^-_t}$ from Kalman filter
    \EndFor
    \State Solve $\mathsf{Q}_t(\Gamma)$ for $\{\mathcal{S}_t,\mathcal{W}_t\}$, and execute $\mathbf{w}_t$
    \For{$h \in \mathcal{H}_{t}^-$}
        \State Predict $v^{h+1,n}_\text{closing}$ and $d^{h+1,n}_\text{prox}$ over $\mathcal{H}^-_t$
    \EndFor
    \State Update $\phi^{h+1,n}=\phi_\text{base}+\phi_\text{proactive}^{h+1,n}$
    \If{$t \bmod T = 0$}
        \State Calculate $L(t)$
        \State Update $\gamma \leftarrow \gamma - \epsilon\frac{\partial L(t)}{\partial\gamma}$ $\forall\gamma \in \Gamma$
    \EndIf
\EndFor
\end{algorithmic}
\end{algorithm}
\vspace{-0.1in}

\section{Experiments}\label{section4}
\vspace{-0.05in}

\begin{figure*}[t]
\centering
    \begin{subfigure}[t]{0.19\textwidth}
      \includegraphics[width=\textwidth]{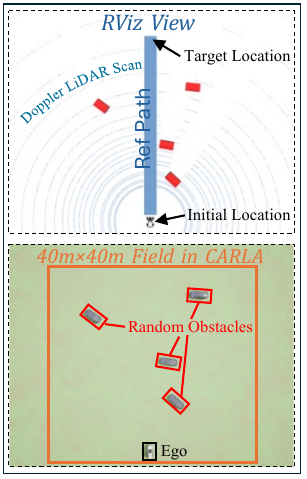}
      \caption{Settings.}
      \label{setting}
    \end{subfigure}
    \begin{subfigure}[t]{0.19\textwidth}
        \includegraphics[width=1.00\textwidth]{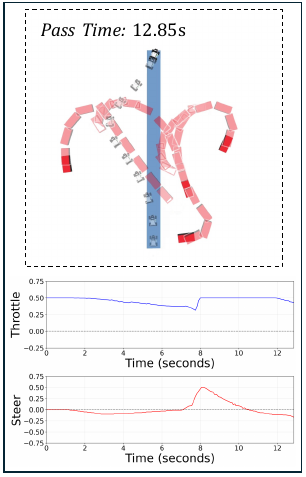}
        \caption{HPTune.}
        \label{hptune}
    \end{subfigure}
    \begin{subfigure}[t]{0.19\textwidth}
        \includegraphics[width=1.00\textwidth]{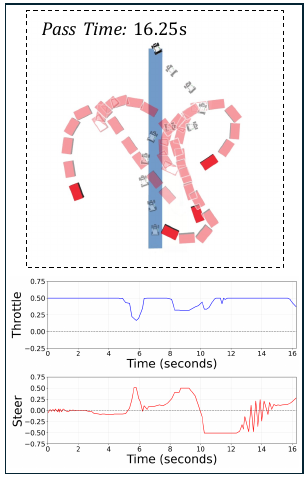}
        \caption{DiffTune-MPC.}
        \label{difftune}
    \end{subfigure}
        \begin{subfigure}[t]{0.19\textwidth}
        \includegraphics[width=1.00\textwidth]{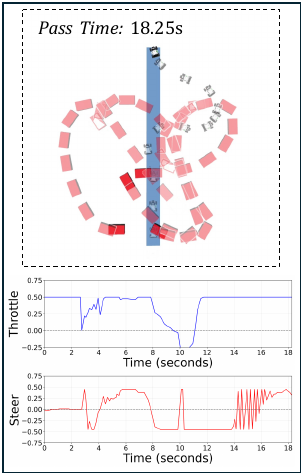}
        \caption{RDA.}
        \label{rda}
    \end{subfigure}
        \begin{subfigure}[t]{0.19\textwidth}
        \includegraphics[width=1.00\textwidth]{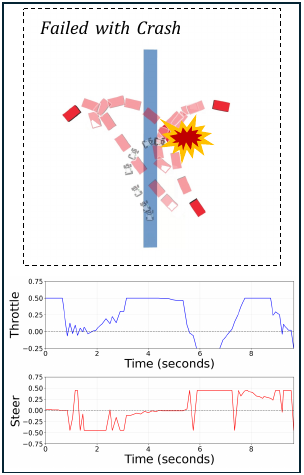}
        \caption{OBCA.}
        \label{obca}
    \end{subfigure}
    \caption{Scene settings and navigation results among 4 random obstacles. (a) shows the experimental scene in CARLA and the corresponding Doppler LiDAR scan view in RViz. (b)-(e) present the ego behaviors for different methods during navigation, where the upper graphs illustrate trajectories and the lower graphs show control actions (throttle and steer) over time.
    }
    \vspace{-0.1in}
\end{figure*}

\begin{figure}[t]
\centering
    \begin{subfigure}[t]{0.23\textwidth}
        \includegraphics[width=1.00\textwidth]{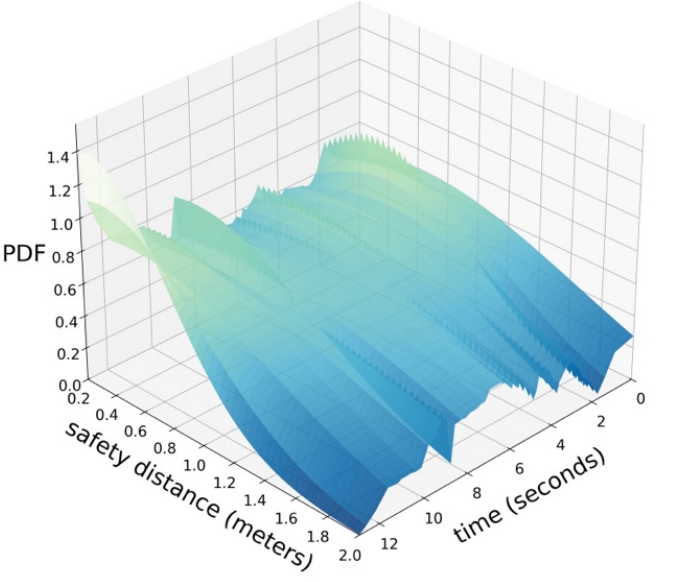}
        \caption{HPTune.}
        \label{hppdf}
    \end{subfigure}
    \begin{subfigure}[t]{0.23\textwidth}
        \includegraphics[width=1.00\textwidth]{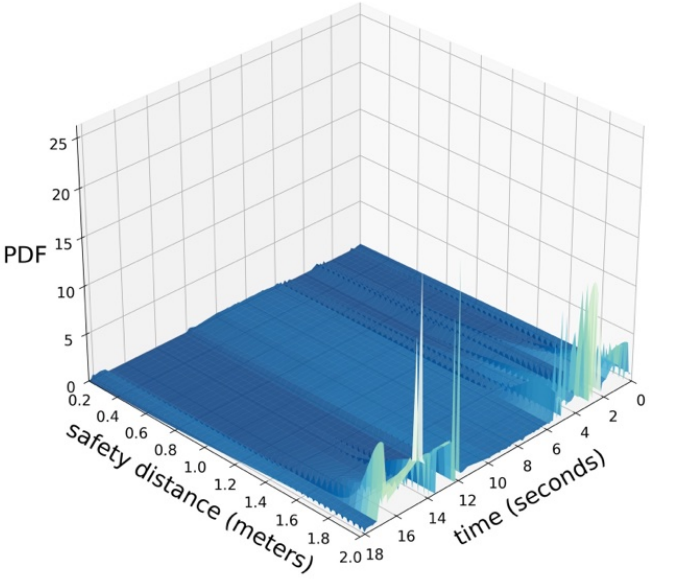}        
        \caption{RDA.}
        \label{rdapdf}
    \end{subfigure}
    \vspace{-0.05in}
    \caption{Run-time PDF of safety distances.}
    \vspace{-0.05in}
    \label{pdf}
\end{figure}

\begin{figure}[t]
\centering
    \begin{subfigure}[t]{0.23\textwidth}
        \includegraphics[width=1.00\textwidth]{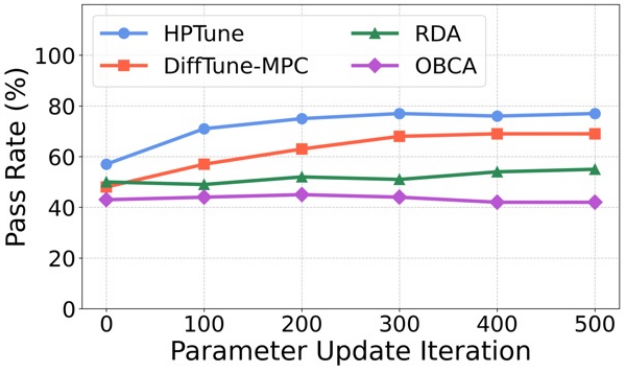}
        \caption{4 random obstacles.}
    \end{subfigure}
    \begin{subfigure}[t]{0.23\textwidth}
        \includegraphics[width=1.00\textwidth]{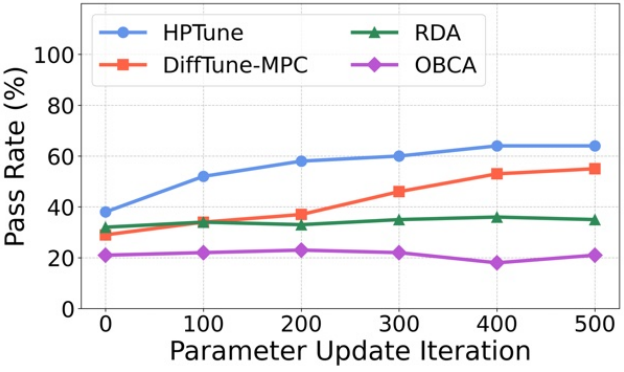}
        \caption{6 random obstacles.}
    \end{subfigure}
    \vspace{-0.05in}
    \caption{Pass rate comparison over parameter update iterations in scenarios with different obstacle densities.}
    \vspace{-0.15in}
    \label{passrate}
\end{figure}

In this section, we perform evaluations by implementing the entire HPTune in high-fidelity simulator CARLA \cite{dosovitskiy2017carla} via ROS bridge. All experiments are conducted on Ubuntu 20.04 system computer with NVIDIA 5060Ti GPU and Intel i7-14700KF CPU, where Pytorch \cite{paszke2019pytorch} is employed to perform backpropagation.
The following methods are compared:
\textbf{HPTune}: our proposed method; 
\textbf{DiffTune-MPC} \cite{tao2024difftune}: a closed-loop MPC tuning method;
\textbf{RDA} \cite{han2023rda}: an open-loop MPC tuning method;
\textbf{OBCA} \cite{zhang2020optimization}: a non-tuning MPC method.

As Fig.~\ref{setting} shows, the experimental scene is in a $40\,\text{m}\times40\,\text{m}$ square field, where both the ego and obstacles are 4-wheel vehicles. Obstacles are spawned at random locations and perform random movements at $5\,\text{m}/\text{s}$ within the square field, while the goal of the ego vehicle is to traverse the area along the reference path by real-time motion planning without any collisions. The hyperparameters are set as follows: $H=15$, $\Delta t=0.12\,\text{s}$ ; $\phi_\text{base}=0.2\,\text{m}$,  $\phi_\text{max}=2.0\,\text{m}$; $\eta_1=10^{-1}$, $\eta_2=10^{-1}$, $\eta_3=10^{-3}$; $T=5$, $\epsilon=10^{-3}$. Tunable parameters are initialized as $\alpha=0.2$ and $\beta=5.0$.

Figs.~\ref{hptune}, \ref{difftune}, \ref{rda}, and \ref{obca} show the navigation results of different methods in a scenario with 4 obstacles generated using the same random seed.
Among these, HPTune demonstrates a superior performance, achieving shorter pass times and smoother control actions compared to the other approaches. 
Fig.~\ref{pdf} depicts the run-time probability density function (PDF) of tuned safety distances during navigation. 
As Fig.~\ref{rdapdf} shows, the open-loop tuning method RDA \cite{han2023rda} adjusts safety distances only reactively when obstacles are nearby, specifically around $t=2\!-\!5~\text{s}$ and $t=12\!-\!17~\text{s}$, which leads to unsmooth control actions in Fig.~\ref{rda}. 
In contrast, HPTune continuously updates $\phi^{h+1,n}$ in a proactive manner throughout the entire run as shown in Fig.~\ref{hppdf}, resulting in smooth, agile, and situation-tailored navigation in Fig.~\ref{hptune}.

To further evaluate the effectiveness and efficiency of HPTune, we compare the pass rate across different parameter update iterations under varying obstacle densities. Specifically, as Fig.~\ref{passrate} presents, for each density level (4 and 6 random obstacles), we measure the pass rate by conducting 50 random trials at every 100-iteration interval during the parameter update process. Results show that HPTune consistently maintains the highest pass rate across all methods and obstacle densities, with improvements of $7\,\%\!-\!46\,\%$.
Particularly, compared to the closed-loop tuning counterpart DiffTune-MPC \cite{tao2024difftune}, HPTune improves pass rate by $13.7\,\%$ and $8.3\,\%$ among 4 and 6 random obstacles, respectively, which confirms the superiority of extended evaluations.
Besides, HPTune shows an approximate convergence on pass rate at about 200 iterations, highlighting its ultra-efficient tuning ability as a result of proactive designs. 

Additionally, we evaluate each method using following key performance metrics: average acceleration (\textbf{AvgAcc}), average jerk (\textbf{AvgJerk}), and average pass time (\textbf{AvgTime}). These results are obtained by averaging over 100 trials, with parameters updated through 500 iterations among 4 random obstacles. Table~\ref{table} summarizes both tuning features and navigation performance of different methods, where HPTune consistently demonstrates clear advantages in motion smoothness and planning efficiency across all metrics.

\begin{table}[t]
\centering
\caption{Tuning features and navigation performances.}
\vspace{-0.15in}
\label{table}
\begin{center}
\scalebox{0.77}{
\begin{tabular}{ccc@{\hspace{-1.0em}}cccc}
\hline
\multirow{2}{*}{\textbf{Method}}  & \multicolumn{2}{c}{\textbf{Tuned Object}} & \multicolumn{1}{c}{\textbf{AvgAcc}\,$\downarrow$} 
& \multicolumn{1}{c}{\textbf{AvgJerk}\,$\downarrow$} & \multicolumn{1}{c}{\textbf{AvgTime}\,$\downarrow$} \\ \cline{2-3} 

 & $C_t$ & $\phi$  &   $(\text{m}/\text{s}^2)$ & $(\text{m}/\text{s}^3)$   & $(\text{s})$ \\ 
\hline
  OBCA & \ding{55} & \ding{55} & 2.160   & 6.014 & 14.695 \\ 

  RDA    & \ding{55} & \ding{51} & 2.302   &6.518& 19.217\\ 

DiffTune-MPC & \ding{51} & \ding{55} & 2.009   & 6.264 &  15.668 \\ 

HPTune & \ding{51} & \ding{51}  &\textbf{1.994}   &\textbf{5.975} &  \textbf{13.732} \\ \hline
\end{tabular}
}
\end{center}
\vspace{-0.25in}
\end{table}

\section{Conclusion}\label{section5}

This paper introduces HPTune, a framework that transforms the traditional failure-triggered, reactive closed-loop MPC parameter tuning process into a hierarchical, proactive closed-loop approach, enabling anticipation-driven and situation-tailored parameter adaptation. Experimental results demonstrate that, compared to benchmark methods, HPTune increases the navigation pass rate by $7\,\%\!-\!46\,\%$ in complex environments, while also significantly enhancing action smoothness and tuning efficiency. These findings underscore the advantages of incorporating motion predictions to extend the closed-loop tuning to the non-executed action perspective for efficient collision-free MPC motion planning.

\bibliographystyle{IEEEbib}
\bibliography{refs}

\end{document}